\documentclass[conference]{IEEEtran}
\IEEEoverridecommandlockouts

\usepackage{cite}
\usepackage{amsmath,amssymb,amsfonts}
\usepackage{amsthm}
\usepackage{bm}
\usepackage{algorithm}
\usepackage{algpseudocode}
\usepackage{graphicx}
\usepackage{textcomp}
\usepackage{xcolor}
\usepackage{booktabs}
\usepackage{pifont}
\usepackage{enumitem}
\usepackage[utf8]{inputenc}
\usepackage{url}
\usepackage[colorlinks=true,allcolors=blue,breaklinks=true]{hyperref}
\usepackage{makecell}
\def\BibTeX{{\rm B\kern-.05em{\sc i\kern-.025em b}\kern-.08em
    T\kern-.1667em\lower.7ex\hbox{E}\kern-.125emX}}

\newtheorem{theorem}{Theorem}
\newtheorem{corollary}{Corollary}

\newcommand{\cmark}{\ding{51}}
\newcommand{\xmark}{\ding{55}}

\newcommand{\QCFS}{\textsf{QCFS}}
\newcommand{\SpikeGen}{\textsf{SpikeGen}}
\newcommand{\SpikeCount}{\textsf{SpikeCount}}
\newcommand{\SoftResetUpdate}{\textsf{SoftResetUpdate}}

\begin{document}

\title{NeuroFlex: Lossless Element-Level ANN-SNN Co-Execution for Efficient Sparse Inference\\
\thanks{This research work was supported in part by the RISC-V Knowledge Centre of Excellence, sponsored by the Ministry of Electronics and Information Technology (MeitY).}
}

\author{\IEEEauthorblockN{Varun Manjunath}
\IEEEauthorblockA{\textit{Electrical Engineering} \\
\textit{Indian Institute of Technology Madras}\\
Chennai, India \\
ee20b149@smail.iitm.ac.in}
\and
\IEEEauthorblockN{Pranav Ramesh}
\IEEEauthorblockA{\textit{Computer Science and Engineering} \\
\textit{Indian Institute of Technology Madras}\\
Chennai, India \\
cs22b015@smail.iitm.ac.in}
\and
\IEEEauthorblockN{Gopalakrishnan Srinivasan}
\IEEEauthorblockA{\textit{Computer Science and Engineering} \\
\textit{Indian Institute of Technology Madras}\\
Chennai, India \\
sgopal@cse.iitm.ac.in}
}

\maketitle

\begin{abstract}
Sparse DNN accelerators specialize in ANN or SNN execution, leaving energy or latency on the table when workload characteristics vary within a layer. Hybrid accelerator designs that switch modes at layer or tile granularity suffer from low PE utilization since one core type idles whenever the other is active. NeuroFlex is the first accelerator to assign every output element independently to ANN or SNN execution mode with zero accuracy loss. We extend integer-exact ANN-SNN equivalence from layers to individual output elements, thereby enabling mode switching with no conversion error. An offline cost-guided scheduler scores each element by its marginal energy-delay trade-off and packs work across PEs, achieving 97-99\% PE utilization compared to 40-45\% for layer-wise hybrids. NeuroFlex reduces EDP by 57-67\% over a strong ANN-only baseline and delivers up to 2.5$\times$ speedup over a dual-sparse SNN-only baseline. Our cost-guided scheduler improves throughput by 16-19\% over random element assignment across vision, language, and transformer workloads.
\end{abstract}

\begin{IEEEkeywords}
Neuromorphic Computing, Sparse Neural Network Accelerators, Spiking Neural Networks (SNNs), ANN-SNN Conversion, Hybrid ANN-SNN Architecture
\end{IEEEkeywords}

\section{Introduction}

Deep neural network (DNN) inference is increasingly deployed at the edge, where stringent power and latency budgets make computational efficiency the primary design constraint. As model architectures diversify from convolutional networks to transformers to billion-parameter large foundational models, hardware must accommodate workloads with fundamentally different compute profiles, both across and within individual layers~\cite{sze2017efficient}. A dominant strategy for reducing inference cost is to exploit \emph{sparsity}: the observation that pruning, quantization, and activation functions drive large fractions of operands to zero, making many of the multiply-accumulate (or MAC) operations in a dense datapath redundant~\cite{han2016deep_compression}.

The early sparse deep learning accelerators targeted a single sparsity source~\cite{han2016eie, albericio2016cnvlutin}. Later designs exploiting \emph{dual sparsity}, i.e., zeros on both operand sides of a dot product, achieved multiplicative reductions in compute and memory traffic, since only matched non-zero pairs require arithmetic~\cite{parashar2017scnn, SparTen, qin2020sigma, gamma, Spada, deng2021gospa}. All of these accelerators, however, remain committed to MAC-based ANN execution. An alternative computing paradigm exists in spiking neural networks (SNNs), which replace full multiply-accumulate operations with accumulate-only (AC) arithmetic over binary spikes, substantially lowering the energy per operation~\cite{roy2019towards}. This has motivated dual-sparse SNN accelerators such as LoAS~\cite{yin2024loas}. However, a spiking neuron reconstructs a single activation value by accumulating spikes over multiple sequential \emph{timesteps}, so its inference latency scales with the timestep count $T$ that the single-pass ANN execution does not incur~\cite{yin2024loas, zhao2024epha}. This creates a fundamental energy-latency trade-off: ANN execution is fast but power-hungry, while SNN execution is energy-efficient but slow.

\begin{figure}[t]
    \centering
    \includegraphics[width=\columnwidth]{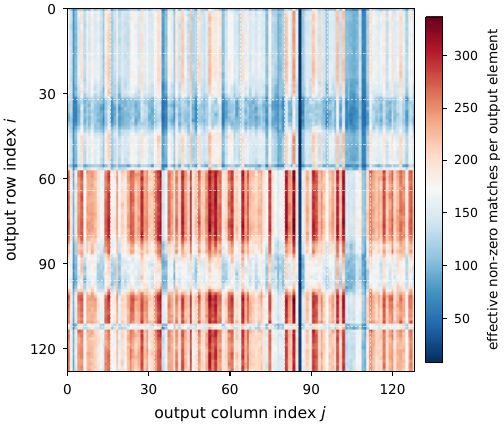}
    \caption{Per-element work varies several-fold within a layer.
    Effective number of operations per output element for VGG-16 conv layer (conv3\_3, $256{\to}256$ channels, $3{\times}3$),
    lowered to a $3136\times256$ GEMM by im2col; the panel shows a
    $128\times128$ crop, and not the full feature map. Each cell is colored by the number
    of operand pairs surviving the bitmap inner-join, the only positions
    requiring arithmetic operation at 63.0\% activation sparsity and 80\%
    globally pruned weights. The remaining work is distributed unevenly
    across the output matrix, with individual $16\times16$ tiles (dashed)
    spanning both high- and low-work elements.}
    \label{fig:element-variation}
\end{figure}

Critically, this trade-off is not uniform across the output elements within a layer. Dual-sparsity execution produces a wide variation in the number of surviving non-zero matches per output element $(i,j)$ of a layer's GEMM or convolution. Figure~\ref{fig:element-variation} makes this concrete for a mid-depth VGG-16 convolution layer: after zero operands on both sides are skipped, the work that remains is spread unevenly across the output matrix, varying several-fold from element to element and structured along both the spatial and the channel dimensions. Elements with many surviving matches are better served by fast ANN execution where the single-pass MAC completes quickly, while elements with few matches favor energy-efficient SNN execution, where fewer operations must amortize the per-element timestep overhead. This per-element variation is the core opportunity that current accelerators leave unexploited, because all existing designs commit to a single execution mode for all elements.

\begin{figure*}[ht]
    \centering
    \includegraphics[clip,width=\textwidth]{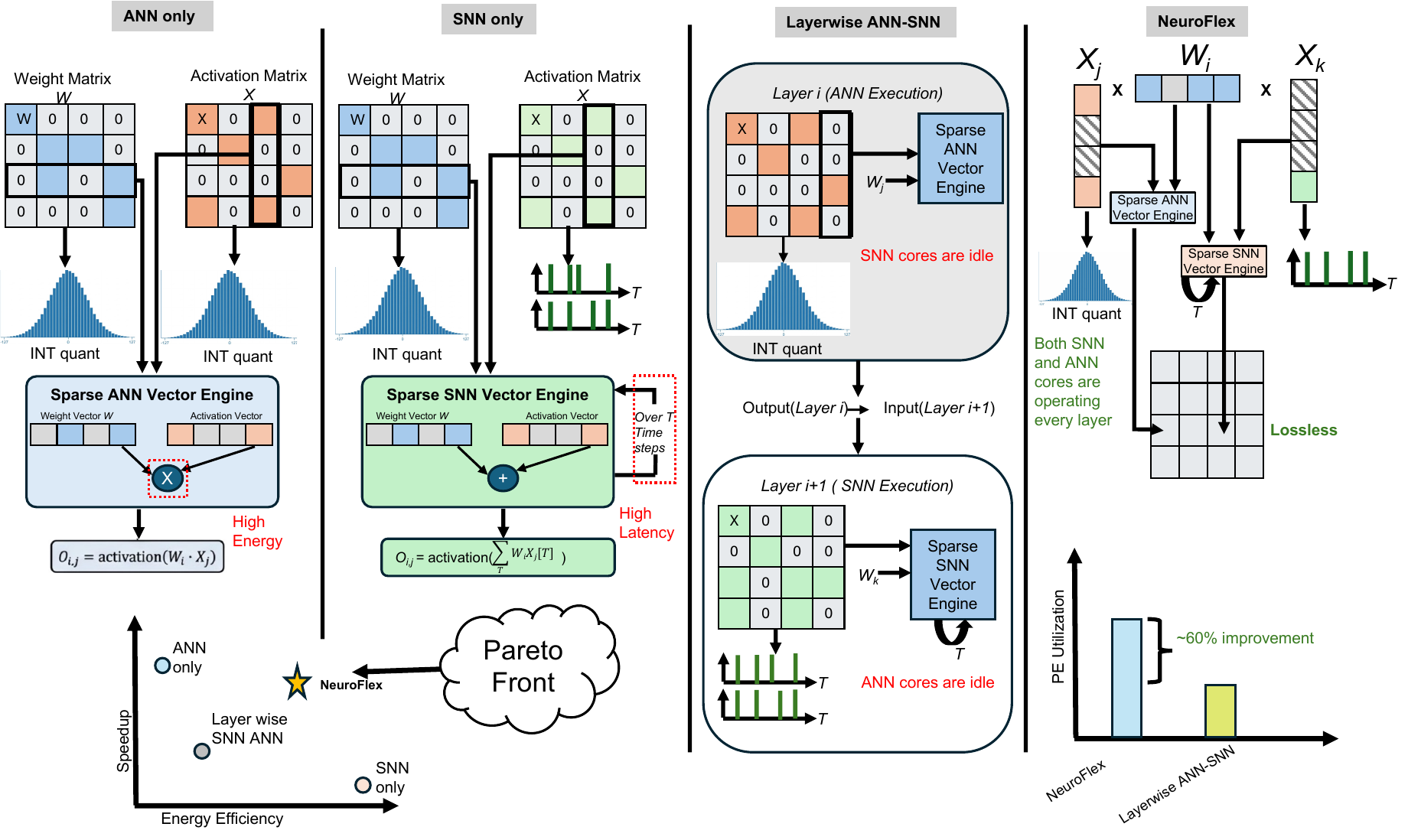}
    \caption{Comparison of execution granularity across accelerators. ANN-only execution is fast but power hungry; SNN-only is energy-efficient but slow. Layerwise ANN-SNN hybrids leave one core type idle per layer, limiting PE utilization. NeuroFlex assigns each output element independently to ANN or SNN execution based on its cost, keeping both cores active every layer and achieving ~40\% higher PE utilization. Empty cells represent zeros.}
    \label{fig:granularity_comparison}
\end{figure*}

Existing hybrid ANN-SNN accelerators attempt to capture this trade-off but operate at coarser granularity. Layer-level designs~\cite{singh2020nebula, zhao2024epha} switch entire layers between ANN and SNN execution modes, thereby idling one core type whenever the other is active. Tile-level designs~\cite{cdnn, c-transformer} improve on this by partitioning layers into fixed-size tiles routed independently, but a tile spanning both high- and low-match elements must still commit to a single mode, leaving intra-tile variation unexploited (Figure~\ref{fig:element-variation}). Moreover, the finest-grained of these designs relies on a lossy ANN-to-SNN conversion, so every tile routed to SNN execution trades accuracy for energy. This accuracy-efficiency coupling has been the central barrier to finer-grained hybrid execution.

The barrier exists because ANN-SNN conversion, which is the technique of training a standard ANN and then algorithmically transforming it into an SNN with matching weights, has historically been approximate. Prior methods~\cite{sengupta2019going, diehl2015fast, bu2023qcfs} calibrate thresholds or firing rates to minimize, but not eliminate, the conversion error. Coarse, layer-level or model-level calibration can average this error out across many neurons. However, an individual output element has no such averaging to hide behind, which is why routing individual elements to SNN execution has been considered unsafe: a misconverted element degrades the accuracy with no mechanism to bound the error. PASCAL~\cite{pascal} recently eliminated this barrier by proving a mathematically exact equivalence between an ANN with Quantization-Clip-Floor-Shift (QCFS) activation and its converted SNN, reproducing the ANN activation under integer arithmetic with zero error rather than a bounded approximation. 
We make a key observation that the PASCAL proof depends solely on \emph{local} integer arithmetic per output element. Membrane potential updates, threshold comparisons, and reset operations are independent across output positions. This means that the equivalence holds at the level of an individual output element $(i,j)$, the finest granularity at which an inner-product dataflow can independently assign work. Hence, any element can execute as ANN or SNN and produce bit-identical results, with no accuracy loss from mixing execution modes within a layer. This observation unlocks lossless element-level hybrid execution.

NeuroFlex is the first accelerator to exploit this element-level equivalence (see Figure~\ref{fig:granularity_comparison}). An offline cost-guided scheduler scores each element by its marginal energy-delay cost, derived from the number of effective non-zero matches and per-core energy and latency coefficients obtained from RTL microbenchmarks. The scheduler uses simulated annealing with element-flip moves to minimize the energy-delay product (EDP), initializing each element on the core with a lower marginal cost, and iteratively refining the assignment to balance load across PEs. Both core types share a unified bitmap-encoded FiberCache and memory hierarchy. On-the-fly spike generation keeps spikes as one-bit binary values by absorbing threshold scaling into the weights at compile time, so the data format is INT8 throughout and mode switching changes only the PE control path, not storage or memory traffic.

Evaluated across CNNs, transformers, and large language models, NeuroFlex reduces EDP by 57-67\% over SparTen~\cite{SparTen}, delivers up to 2.5$\times$ speedup over LoAS~\cite{yin2024loas}, and achieves 97-99\% PE utilization, compared to 40-45\% for layer-wise and 60-83\% for tile-wise hybrid baselines in our evaluation. The cost-guided scheduler improves throughput by 16-19\% over random element assignment, quantifying the value of per-element energy-delay matching. In summary, this paper makes the following contributions:

\begin{itemize}
    \item We prove that PASCAL's integer-exact ANN-SNN equivalence extends to individual output elements, enabling lossless element-level mode switching within a layer.
    \item We propose a cost-guided offline scheduler that utilizes simulated annealing to assign each element to the core that minimizes its marginal energy-delay cost, achieving 97-99\% PE utilization.
    \item We design NeuroFlex, a heterogeneous accelerator with dual ANN and SNN core types that co-execute at element granularity, the finest level at which the inner-product dataflow can independently assign work.
    \item We evaluate NeuroFlex across diverse vision, language, and transformer models, demonstrating substantial gains in EDP, speedup, energy efficiency, and PE utilization over both single-mode and coarse-grained hybrid baselines.
\end{itemize}
\section{Related Work}
\label{sec:related}

\subsection{ANN-SNN Conversion}
Converting a trained ANN into an SNN requires reconciling two different computational primitives: ANNs compute a real-valued activation per neuron, while SNNs communicate only through binary spikes over a sequence of timesteps. QCFS\cite{bu2023qcfs} closes this gap at training time by quantizing each ANN activation into $L$ discrete levels, so that the activation's value is already expressible as a spike count out of $L$ possible firing opportunities. After training, each QCFS-activated neuron is converted into a spiking neuron that runs for a fixed number of phases and emits spikes whose cumulative count reproduces the QCFS output. PASCAL~\cite{pascal} shows that, for integer-valued inputs, this conversion is exact rather than approximate: the spiking neuron's output is bit-identical to the QCFS activation it replaces. We detail the resulting three-phase spiking neuron below and refer readers to~\cite{pascal} for the full derivation.

Early ANN-SNN conversion methods map ReLU activations to firing rates of Integrate and Fire (IF) neurons in order to calibrate thresholds or weights and limit conversion error. This enables deep SNNs on CIFAR/ImageNet, while typically requiring longer timesteps and floating-point calibration~\cite{sengupta2019going}. More recent work derives conversion error terms and replaces ReLU with Quantization-Clip-Floor-Shift (QCFS) to reduce latency while preserving accuracy using fewer timesteps~\cite{bu2023optimal}. However, these conversion methods still suffer from conversion inaccuracies, especially for larger datasets. PASCAL~\cite{pascal} instead proves a mathematically exact equivalence between an ANN with QCFS activation and its converted SNN with spike inhibition and accumulation per-neuron, yielding lossless integer execution.

The QCFS function, $\widehat{h}$, as proposed by~\cite{bu2023optimal}, can be formulated as
\begin{align}
    \label{annnew2}
        \widehat{h}(\bm{z}^{l})=\lambda^l~\mathrm{clip} \left(\frac{1}{L}\left \lfloor \frac{\bm{z}^{l}L}{\lambda^l}+ \frac{1}{2} \right \rfloor, 0, 1 \right),
\end{align}
where $z^l$ is the input to the activation layer, $L$ is the quantization step, and $\lambda^l$ is the trainable parameter for the activation threshold. PASCAL~\cite{pascal} uses the following algorithmic approach for precise ANN-SNN conversion using a three-step process. For a given threshold $\lambda_l$ and a quantization step $L$, threshold $\theta$ is initialized to $\frac{\lambda_l}{L}$, and the initial membrane potential $V_0 = \frac{\theta}{2}$. The three phases in the PASCAL spiking neuronal layer are summarized below.

\begin{itemize}
    \item \textbf{Phase 1: Accumulation ($t = 0, \ldots, L - 1$)} ,  Integrate input and generate spikes through standard integrate-and-fire operation:
    \begin{align}
    V_0^{(1)} &= 0.5\theta \\
    V_t^{(1)} &= V_{t-1}^{(1)} + x[t] - H(V_{t-1}^{(1)} + x[t] - \theta) \cdot \theta \\
    C_1 & = \sum_{t=0}^{L-1} H(V_{t-1}^{(1)} + x[t]) \cdot \theta
    \end{align}

    \item \textbf{Phase 2: Excitatory and Inhibitory Spiking ($t = L, \ldots, 2L - 2$)} ,  Generate both excitatory and inhibitory spikes based on accumulated potential:
    \begin{align}
    V_{L-1}^{(2)} &= V_{L-1}^{(1)} \\
    V_t^{(2)} &= V_{t-1}^{(2)} - H(V_{t-1}^{(2)} - \theta) \cdot \theta + H(-V_{t-1}^{(2)}) \cdot \theta \\
    C_2 &= \sum_{t=L}^{2L-2} H(V_{t-1}^{(2)} - \theta) \cdot \theta - H(-V_{t-1}^{(2)}) \cdot \theta
    \end{align}

    \item \textbf{Phase 3: Output Generation ($t = 2L - 1, \ldots, 3L - 2$)} ,  Sum excitatory and inhibitory contributions and generate final output spikes:
    \begin{align}
    V_{2L-2}^{(3)} &= C_1 + C_2 \\
    V_t^{(3)} &= V_{t-1}^{(3)} - H(V_{t-1}^{(3)} - \theta) \cdot \theta
    \end{align}
\end{itemize}
The PASCAL algorithm~\cite{pascal} proves that for integer-valued inputs, the QCFS activation can be exactly reproduced by a three-phase integrate-fire-reset sequence using $(3L{-}1)$ iterations. The equivalence can be stated compactly as:
\begin{equation}
  h_b(z^l) \;=\; \Sigma_{t=2L-1}^{3L-2} V_t^{(3)},
  \label{eq:pascal-compact}
\end{equation}
This key property holds for every element, guaranteeing zero conversion error under integer arithmetic. NeuroFlex exploits the per-element equivalence to enable element-level mode switching with no accuracy loss.

\begin{table}[h]
\centering
\caption{Notation for PASCAL spiking neuron model.}
\label{tab:notation}
\begin{tabular}{l|l}
\hline
\textbf{Notation} & \textbf{Description} \\
\hline
$V_t^{(i)}$ & Membrane potential at time $t$ in phase $i$ \\
$x[t]$ & Input signal at time $t$ \\
$H(\cdot)$ & Heaviside step function \\
$\theta$ & Firing threshold \\
$C_1$ & Cumulative spike count from Phase 1 \\
$C_2$ & Net excitatory-inhibitory spike count from Phase 2 \\
$L$ & Duration of each phase in timesteps \\
\hline
\end{tabular}
\end{table}

\subsection{Sparsity Exploitation in DNN Accelerators}

Pruning, quantization, and activation functions drive substantial fractions of both weights and activations to zero, thus creating dual sparsity, i.e., the simultaneous presence of zeros on both operand sides of a dot product~\cite{SparTen, han2016eie}. Exploiting dual sparsity yields multiplicative reductions in compute and memory traffic because only matched non-zero pairs require computation. However, the density of matched pairs varies widely across output elements within the same layer: some elements have many surviving matches (dense dot products) while others have very few (sparse dot products). This per-element variation is the core opportunity targeted by NeuroFlex.

Early sparse accelerators exploited a single sparsity source \cite{han2016eie, albericio2016cnvlutin}. The shift to dual sparsity has since driven larger gains across a range of dataflows and encodings~\cite{parashar2017scnn,qin2020sigma, Spada, gamma}. SparTen~\cite{SparTen} proposed a bitmap-based inner-join design that intersects activation and weight bitmaps to identify matched non-zero pairs, with software-level load balancing for unstructured sparsity. Bitmap encoding incurs minimal metadata overhead, one bit per element, keeping memory traffic low. Gamma~\cite{Spada} accelerates sparse matrix multiplication using Gustavson's algorithm with efficient merge operations, but relies on CSR-style coordinate encoding whose multi-byte metadata inflates memory traffic relative to bitmap-based designs. All of these accelerators, however, are designed exclusively for ANN computation: the MAC unit remains the fundamental compute primitive, with no mechanism to exploit the cheaper accumulate-only arithmetic available in SNN execution.

On the SNN side, dual sparsity arises from binary spikes and pruned weights, but the temporal dimension introduces an extra loop into the Sparse Matrix{-}Sparse Matrix (spMspM) computation that increases latency and memory traffic. LoAS~\cite{yin2024loas} addresses this with a fully temporal-parallel (FTP) dataflow that parallelizes all timesteps and a spike compression mechanism that packs binary spikes contiguously in memory. Despite the temporal parallelism, LoAS's pipeline retains per-chunk overhead from its laggy prefix-sum circuit (8 cycles per chunk) and correction accumulators needed to reconcile fast and laggy prefix outputs. These overheads render SNN execution energy-efficient but inherently slower than ANN execution, particularly for low-sparsity columns where temporal processing cost is not offset by sufficient zero-skipping. As a single-mode SNN-only design, LoAS cannot route such columns to faster MAC-based execution, a limitation shared symmetrically by ANN-only accelerators that cannot exploit SNN energy savings. Neither class can dynamically assign columns to the more cost-effective mode, which is the gap addressed by NeuroFlex.

\subsection{Hybrid ANN-SNN Algorithms and Hardware Architectures}

On the algorithm side, hybrid ANN-SNN models attempt to combine the accuracy and speed of ANNs with the energy efficiency of SNNs within a single inference pipeline. Aydin et al.~\cite{aydin2024hybrid} proposed a ``slow-fast'' scheme where an auxiliary ANN running at low rate initializes SNN membrane potentials, mitigating the long transient periods and state decay that deteriorate standalone SNN accuracy. Layer-wise hybrid algorithms go further by interleaving ANN and SNN blocks within the same model: recent surrogate-gradient formulations enable differentiable encode-decode SNN blocks embedded in ANN pipelines~\cite{luu2025hybrid}. These algorithmic advances establish that mixing ANN and SNN computation can improve the energy-latency trade-off. However, they leave the hardware mapping problem open, specifically, how to keep both ANN and SNN compute resources busy when executing a hybrid workload.

On the hardware side, existing hybrid accelerators operate at coarse granularity. NEBULA~\cite{singh2020nebula} demonstrated a spintronic substrate that supports SNNs, ANNs, and hybrids on a single chip, but switches between modes at the layer-level without exploiting sparsity in either domain. EPHA~\cite{zhao2024epha} provides a parallel hybrid accelerator with configurable ANN/SNN modes and reports energy-latency trade-offs when partitioning models between modes, but again operates at layer granularity and does not incorporate sparsity-aware dataflows. Layer-level switching leaves one core type idle whenever the other is active, resulting in low PE utilization. Our experiments show that layer-wise hybridization achieves only 40-45\% utilization across the evaluated models.

C-DNN~\cite{cdnn} moves to a finer granularity by partitioning workloads across heterogeneous ANN and SNN cores at the tile-level, using a workload allocator that routes tiles to the more energy-efficient domain based on a magnitude-energy trade-off. However, C-DNN's ANN-to-SNN conversion is lossy, i.e., the SNN accuracy is lower than the source ANN, which means that routing tiles to the SNN core trades accuracy for energy. Additionally, tile-level partitioning still leaves significant scheduling slack: tiles are coarser than individual output columns, and hence, the allocator cannot exploit fine-grained variation in sparsity and computational cost within a tile, leading to suboptimal PE utilization. C-DNN also does not exploit dual sparsity in either its ANN or SNN cores.

NeuroFlex addresses all three limitations. First, by building on PASCAL's integer-exact conversion, element-level mode switching introduces zero accuracy loss, and hence, ANN and SNN execution are mathematically equivalent realizations of the same computation. Second, column granularity is the finest level at which the inner-product dataflow can independently assign work, enabling the scheduler to match each column's sparsity characteristics to the most cost-effective execution mode. Third, both cores exploit dual sparsity through bitmap-based inner-join and compressed FiberCache, ensuring that the energy advantage of SNN execution is not offset by wasted cycles on zero operands. Table~\ref{tab:positioning} summarizes the distinctions offered by NeuroFlex compared to prior works.
\begin{table}[t]
\centering
\caption{Comparison of NeuroFlex with prior accelerators.}
\label{tab:positioning}
\resizebox{0.5\textwidth}{!}{%
\begin{tabular}{lccccc}
\toprule
\textbf{Property} & \textbf{SparTen \cite{SparTen}} & \textbf{LoAS \cite{yin2024loas}} & \makecell{\textbf{NEBULA \cite{singh2020nebula}} \\ \textbf{EPHA \cite{zhao2024epha}}} & \makecell{\textbf{C-DNN \cite{cdnn}} \\ \textbf{C-Transformer \cite{c-transformer}}} & \textbf{NeuroFlex} \\
\midrule
Fast             & \cmark & \xmark  & \xmark & \cmark & \cmark \\
Energy Efficient              & \xmark  & \cmark & \cmark & \cmark & \cmark \\
Dual-sparsity     & \cmark & \cmark & \xmark  & \xmark  & \cmark \\
Lossless ANN $\Longleftrightarrow$ SNN  & \cmark & \cmark & \xmark & \xmark & \cmark \\
\bottomrule
\end{tabular}%
}
\end{table}

\section{Lossless ANN-SNN Conversion and Dataflow}
\label{sec:conversion}
We build on the PASCAL framework, which demonstrates a mathematically exact mapping between an ANN with QCFS activation and its converted SNN counterpart. PASCAL achieves this by augmenting the integrate-and-fire (IF) neuron with spike accumulation and inhibition such that the IF process exactly reproduces the ANN activation statistics. The mathematical equivalence was proved at layer-wise granularity and shown to hold under integer arithmetic without approximation.

\subsection{Element-level Extension of PASCAL Equivalence}

The element-level equivalence, as specified by PASCAL for our architecture, is stated in the form of the following corollaries.
\begin{corollary}
A model with quantization step $L_n$ and threshold $\theta_n$ such that $\frac{\theta_n}{2L_n}$ \textbf{is an integer and} $\theta_n$  \textbf{is within the range of INT8} in the input layer satisfies the input layer equivalence as stated in Theorem 3.5 in~\cite{pascal}.
\label{corollary1}
\end{corollary}

\begin{corollary}
A layer with quantization step $L$, performing a MatMul operation followed by Batch Normalization, i.e. $z^{l+1} = \gamma \cdot \frac{(z^l * W) + b - \mu}{\sqrt{\sigma^2 + \epsilon}} + \beta $ such that \textbf{each of the quantities $b' = \frac{b}{L}, \mu' = \frac{\mu}{L}, \beta' = \frac{\beta}{L}$ and $\frac{\gamma}{\sqrt{\sigma^2 + \epsilon}}$ are integers within the limits of INT8} satisfies Theorem 3.3 in~\cite{pascal}.
\label{corollary2}
\end{corollary}
NeuroFlex extends the PASCAL formulation from layer-level to \textit{element-level} hybrid execution. Since the PASCAL proof of correctness depends only on local integer arithmetic and quantization parameters, the same reasoning applies to each element independently.

\begin{theorem}
 For any output element $(i,j)$ operating with integer threshold $\theta$ and quantization step $L$, the SNN realization $M_{i,j}$ is mathematically equivalent to its ANN realization $N_{i,j}$, i.e.,
\[
f_z^{M_{i,j}} = f_z^{N_{i,j}}, \quad \forall z.
\]
\end{theorem}
\begin{figure}[tbp]
  \centering
  \includegraphics[width=0.5\textwidth]{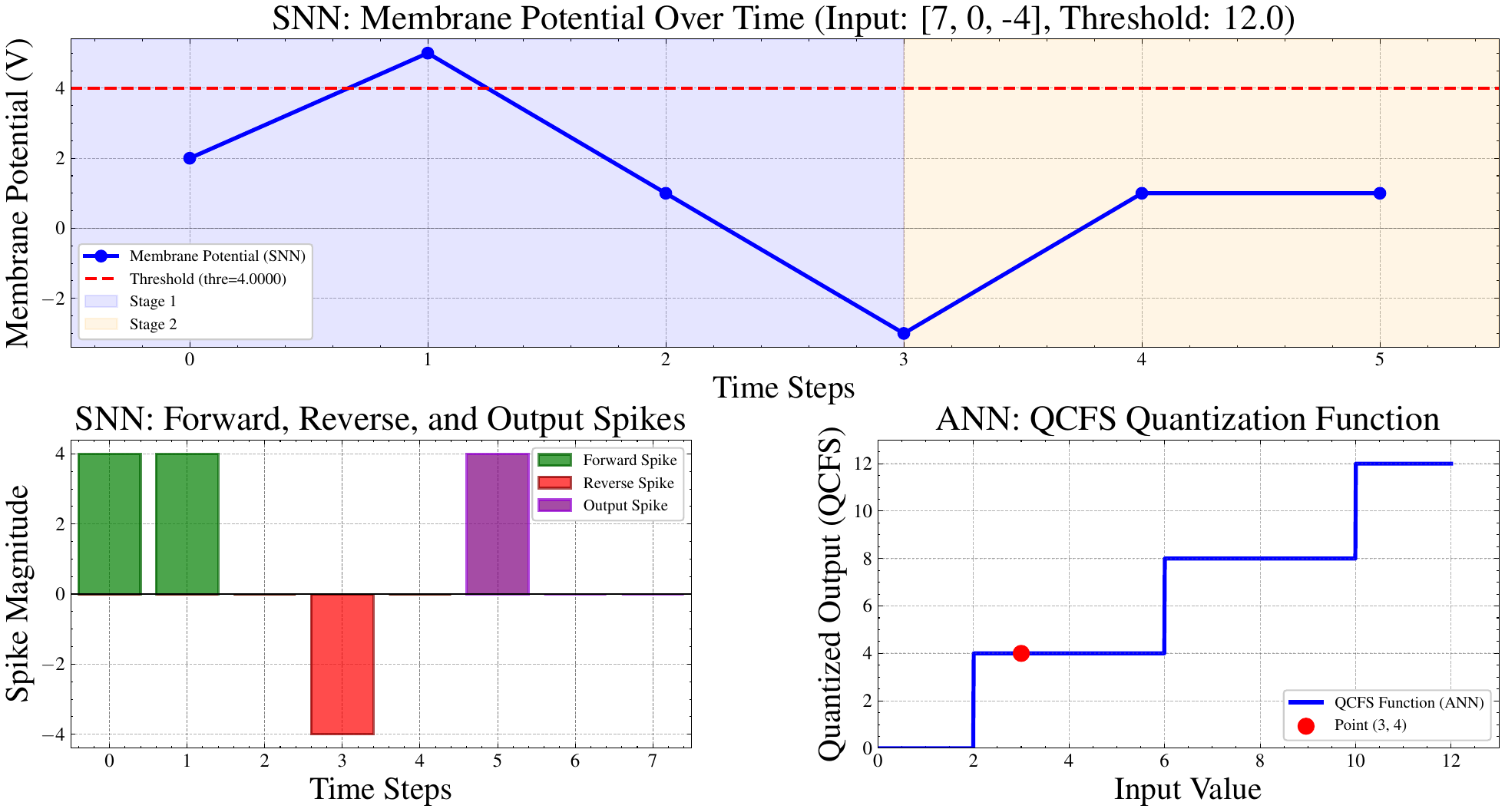}
  \caption{Membrane potential dynamics of a spiking neuron over time. The plot illustrates the temporal evolution of the membrane voltage with a threshold of 12, $L = 3$, input values of [7, 0, -4]. The top line plot shows the first two stages of the PASCAL algorithm, while the bar chart below shows the nature of spikes emitted (forward, inhibitory, and output). This procedure can be shown to be equivalent to applying the QCFS activation to a matrix whose values are the sum of the values ($7 + 0 + (-4)  = 3$) of the input matrices across timesteps.}
  \label{fig:snn-ann}
\end{figure}
\noindent\textbf{Sketch of proof.}
The result follows from Theorems 3.3-3.6 in~\cite{pascal}. The integrate-fire accumulation, thresholding, and reset operations are local to each output element under integer arithmetic. Membrane potential updates depend only on the input activations and weights contributing to that element, and are independent of other output elements. Therefore, temporal accumulation over timesteps reproduces the corresponding ANN activation exactly for each $(i,j)$. Figure~\ref{fig:snn-ann} illustrates this process for a concrete example with threshold 12, $L=3$, and inputs $[7, 0, -4]$.

\subsection{Unified INT8 Dataflow and On-the-Fly Mode Switching}
\label{sec:dataflow-mode}

All activations and weights in NeuroFlex are stored exclusively in INT8 format. No metadata distinguishes ANN data from SNN data; that is, the execution mode is a property of the \emph{schedule}, not the \emph{data}. The offline scheduler produces a per-element binary mask(mode); the PE handles any necessary conversion internally, and the intermediate spike representation is ephemeral, never written to memory. Switching an element's mode therefore changes only the PE's control path, not the storage format or memory traffic.

In the ANN mode, the PE performs a MAC over matched non-zero pairs followed by QCFS activation and writes the INT8 result to SRAM. In SNN mode, the PE converts each matched non-zero INT8 activation into a binary spike pattern via combinational logic, accumulates weighted spikes with threshold comparison, and resolves residual membrane charge, corresponding to the three PASCAL phases. These phases require $3L{-}1 = 23$ sequential iterations for $L{=}8$ ($L$ for spike generation, $L{-}1$ for Spike count, $L$ for membrane potential reinitialization), where each iteration depends on the previous membrane state. This iteration count indicates a \emph{sequential dependency chain in the algorithm}, not an increase in data width, the data remains 8-bit throughout. The output is written back in INT8, identical in format to the ANN path. Both modes follow an inner-product dataflow (Algorithm~\ref{alg:ip-dataflow}), sharing the same FiberCache access pattern and operand reuse schedule. Mixed-mode execution within a layer introduces no additional memory traffic.

\begin{algorithm}
\caption{Inner-Product Dataflow with Column-Wise ANN or SNN Execution.}
\label{alg:ip-dataflow}
\begin{algorithmic}[1]
  \State \textbf{Require:} Input activations $A\in\mathbb{Z}^{M\times K}$ with bitmap $U\in\{0,1\}^{M\times K}$,
  \State \hspace*{1.8em}weights $B\in\mathbb{Z}^{K\times N}$ with bitmap $X\in\{0,1\}^{K\times N}$,
  \State \hspace*{1.8em} $\mathsf{mode}\in\{\mathsf{ANN},\mathsf{SNN}\}^{M\times N}$ (per element),
  \State \hspace*{1.8em}quantization step $L$, spike window (timesteps) $T=L$
  \State \textbf{Ensure:} Output activations $C\in\mathbb{Z}^{M\times N}$
  \State $C \gets 0$
  \For{$m=1$ \textbf{ to } $M$}
    \State \hspace*{0.3em} \textcolor{blue}{// row (output) index}
    \For{$n=1$ \textbf{ to } $N$}
      \State \hspace*{0.3em} \textcolor{blue}{// element index; scheduled as ANN or SNN}
      \If{$\mathsf{mode}[m,n]==\mathsf{ANN}$}
        \For{$k=1$ \textbf{ to } $K$}
          \State \hspace*{0.3em} \textcolor{blue}{// inner-product MAC with dual sparsity}
          \If{$U[m,k]=1$ \textbf{ and } $X[k,n]=1$}
            \State $C[m,n]\gets C[m,n] + A[m,k]\cdot B[k,n]$
          \EndIf
        \EndFor
        \State $C[m,n]\gets \QCFS\!\big(C[m,n]\big)$
      \Else
        \State \hspace*{0.3em} \textcolor{blue}{// SNN path (lossless via PASCAL)}
        \State $O[1{:}T]\gets 0$ \hspace*{0.5em} \textcolor{blue}{// timestep buffer (conceptual)}
        \For{$k=1$ \textbf{ to } $K$}
          \State \hspace*{0.3em} \textcolor{blue}{// selective spike regeneration after match}
          \If{$U[m,k]=1$ \textbf{ and } $X[k,n]=1$}
            \State \hspace*{0.3em} \textcolor{blue}{// only matched positions}
            \State $\mathbf{s}{\gets}\SpikeGen\!\big(A[m,k],L\big)$\hspace*{0.1em}\textcolor{blue}{// $\mathbf{s}{\in}\{0,1\}^{T}$}
            \For{$t=1$ \textbf{ to } $T$}
              \State \hspace*{0em} \textcolor{blue}{// accumulation spatially unrolled $t$}
              \State $O[t]\gets O[t] + \mathbf{s}[t]\cdot B[k,n]$ \hspace*{0.1em}
            \EndFor
          \EndIf
        \EndFor
        \For{$t=1$ \textbf{ to } $T$}
          \State $C[m,n]\gets \SpikeCount\!\big(C[m,n],O[t]\big)$
        \EndFor
        \State $C[m,n]\gets \SoftResetUpdate\!\big(C[m,n]\big)$
      \EndIf
    \EndFor
  \EndFor
\end{algorithmic}
\end{algorithm}

\section{NeuroFlex Architecture}
\label{sec:arch}

\subsection{System Overview}
\begin{figure}[t]
    \centering
    \includegraphics[width=0.9\linewidth]{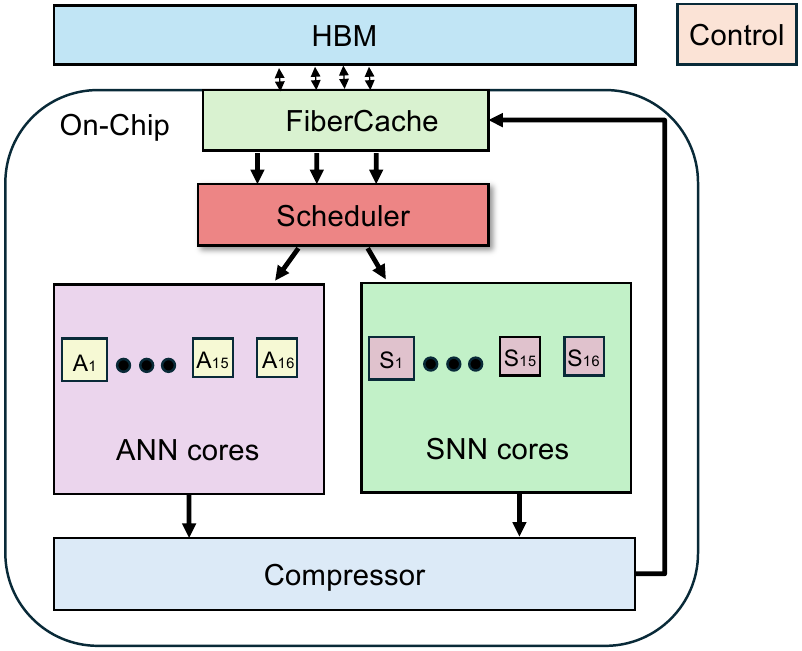}
    \caption{Top-level organization of the proposed NeuroFlex accelerator showing unified memory, shared compressor, FiberCache, and the dual ANN and SNN compute cores.}
    \label{fig:entirechip}
\end{figure}

NeuroFlex integrates heterogeneous ANN and SNN compute cores within a shared memory hierarchy, as shown in Figure~\ref{fig:entirechip}. Each core contains an array of processing elements (PEs) that execute columns dispatched by a centralized scheduler. Both cores draw data from a unified FiberCache backed by HBM, and all outputs pass through a shared compressor before write-back to on-chip SRAM. This organization allows ANN and SNN columns to execute concurrently on their respective cores and share the same storage and data movement infrastructure.

\subsection{Core Organization}
Each core contains multiple PEs that operate on independent vectors streamed from the shared FiberCache unit. The ANN core performs MAC operations followed by QCFS activation. The SNN core executes the PASCAL equivalent integrate-fire sequence using accumulate-only arithmetic. Within a core, idle PEs are filled greedily to maintain load balance.

\begin{figure}[t]
    \centering
    \includegraphics[width=0.70\linewidth, height=0.80\linewidth, clip]{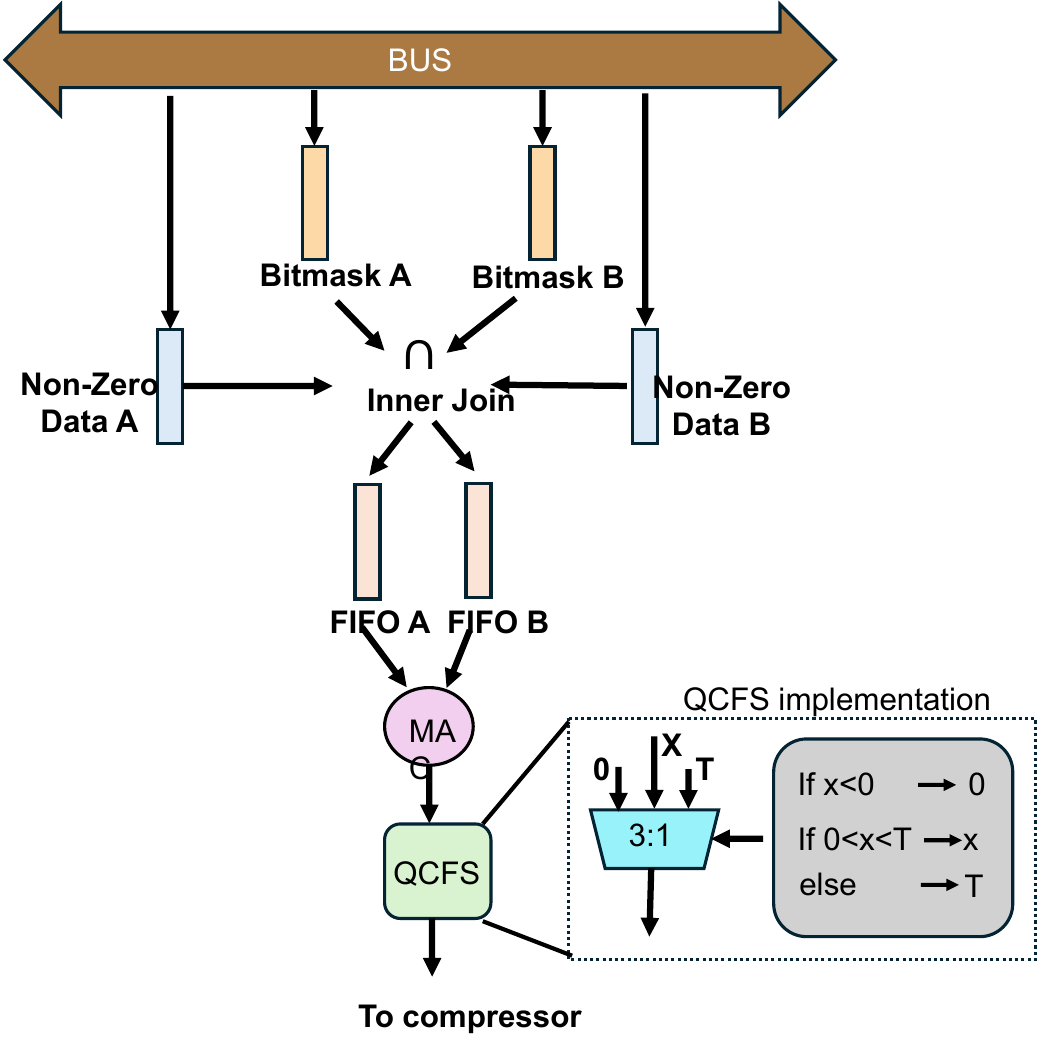}
    \caption{Organization of the ANN core, where each PE performs fetch, prefix alignment, MAC accumulation, and QCFS activation before immediate write-back.}
    \label{fig:anncore}
\end{figure}

\begin{figure}[t]
    \centering
    \includegraphics[width=0.5\textwidth, height=0.4\textwidth, clip]{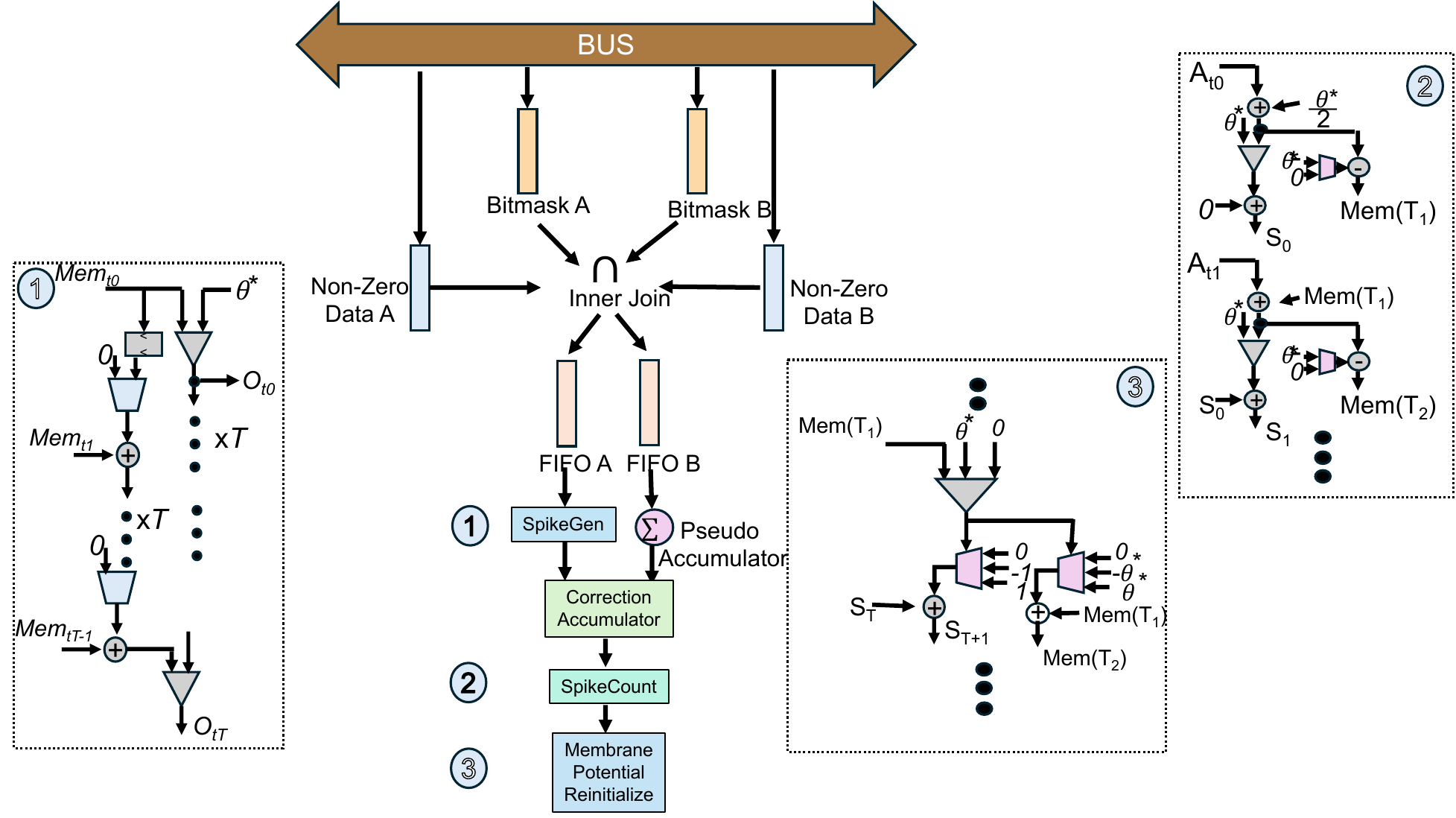}
    \caption{Organization of the SNN core, where each PE includes dual prefix-sum circuits, pseudo-accumulators, and correction accumulators to realize PASCAL-equivalent integrate-and-fire computation. Data~A and B correspond to activation inputs and weights, respectively. Only non-zero activations in Data~A that match active weights undergo spike generation.}
    \label{fig:snncore}
\end{figure}

\subsection{Processing-Element Microarchitecture}
\label{sec:PE}
\paragraph{Dataflow}
Each PE receives a chunk consisting of a bitmap and its corresponding packed non-zero values from the FiberCache. The PE processes one non-zero element per cycle, inserting a single bubble between consecutive chunks to synchronize cache fetches. Both cores maintain identical control sequencing so that workloads can be reassigned transparently between modes.

The \textbf{ANN PE} follows an inner-product dataflow derived from~\cite{SparTen}, as shown in Figure~\ref{fig:anncore}. Upon chunk arrival, a fast prefix circuit~\cite{ParallelPrefix} determines read offsets from the bitmap, and an address generator aligns the corresponding activation and weight streams. The MAC array then accumulates partial sums into a local register tree that overlaps computation across consecutive elements. Once the accumulation for one neuron completes, the QCFS activation stage applies the quantization-clipped nonlinear transform in a single pipeline stage. Under INT configuration where $T = L$ and the quantization step $\lambda = T/L = 1$, the QCFS function simplifies to a saturating clamp: values below zero are clamped to zero, values above $T$ are clamped to $T$, and all others pass through unchanged. This reduces the QCFS hardware to two comparators and a multiplexer. The result is written directly to the on-chip SRAM for reusing it for the next layer's computation.

\textbf{SNN PE} (refer Figure~\ref{fig:snncore}) implements a modified PASCAL dataflow organized into three pipelined stages that together realize the full integrate-fire-reset sequence of the converted SNN neuron. Two prefix-sum circuits operate at the front end to identify active computation regions. A fast prefix produces early offsets for address alignment within a single cycle, while a laggy prefix uses 16 adders and a 128-bit buffer to complete the accumulation of valid matches between activation (Data~A) and weight (Data~B) bitmaps over 8 cycles. Only activations that correspond to matched positions are forwarded downstream, avoiding unnecessary spike encoding for zero or unmatched elements.

\emph{Stage~1, Spike Generation.}
Each matched non-zero INT8 activation is consumed by the spike-generation unit, which deterministically converts it into an $8$-bit binary spike train encoding the same quantized magnitude. The conversion is purely combinational: the activation value determines which of the $8$ temporal slots carry a spike, following the PASCAL encoding scheme. Spike trains are generated selectively, only for activation-weight pairs that survive the inner-join, so the unit is active solely for effective computations. The generated binary spikes are then paired with the corresponding INT8 weight and forwarded to the accumulation stage.

\emph{Stage~2, Spike Count.}
The spike-count stage realizes the excitatory phase of the PASCAL integrate-and-fire neuron. It unrolls $L{-}1$ timesteps in hardware: at each unrolled step, the product of the binary spike and the matched weight is added to the membrane potential register, which is then compared against the firing threshold~$\theta^{*}$. Whenever the membrane meets or exceeds~$\theta^{*}$, an excitatory spike is registered and the membrane is soft-reset by subtracting~$\theta^{*}$. Since the weighted spike inputs are 11-bit signed values and the membrane can accumulate over consecutive timesteps without firing (e.g.\ repeated negative inputs), the membrane registers are provisioned at 14-bit signed width to cover the worst-case range. An unsigned 4-bit counter tracks cumulative excitatory spikes across the $7$ unrolled steps. At the stage boundary, the final membrane value and spike count are latched into pipeline registers for the next stage.

\emph{Stage~3, Membrane Potential Reinitialization.}
The third stage resolves any residual membrane charge that was not fully absorbed during the excitatory phase. It iterates for up to $8$ timesteps with no new external input: if the latched membrane is at or above~$\theta^{*}$, an additional excitatory spike is emitted and $\theta^{*}$ is subtracted; if the membrane is negative, an inhibitory spike is emitted and $\theta^{*}$ is added. This bidirectional correction exactly reproduces the QCFS clip-and-floor non-linearity in the temporal domain and is essential for lossless equivalence when weighted inputs have driven the membrane substantially below zero. Because inhibitory spikes can now decrement the count, the spike-count register widens to 5-bit signed at this stage, while the membrane register retains its 14-bit signed width. This value is further processed through compressor and written back to the memory.

\subsection{Memory Hierarchy and Interconnect}
\label{sec:arch:memory}

Both cores share a unified FiberCache backed by HBM, where each cache line stores a bitmap and a continuation pointer for fibers that span multiple lines. The cache is heavily banked to allow concurrent PE access, enabling overlapping reads of activation fibers and broadcast of compressed weight fibers, with a reuse-based replacement policy~\cite{gamma, Spada} that retains high-reuse fibers. A swizzle-switch crossbar~\cite{crossbar} connects the FiberCache banks to PE clusters, delivering 128-bit bitmap-plus-data chunks per transfer with a single-credit backpressure to avoid stalls. All core outputs pass through a shared compressor that encodes results in bitmap form, one bit of metadata per element, before writing back to on-chip SRAM. The scheduler interfaces with both cores through a unified command queue, issuing a pre-computed mode token (\texttt{ANN} or \texttt{SNN}) and PE assignment per element based on the offline cost function (Section~\ref{sec:scheduling}).

\section{Cost Function and Scheduling}
\label{sec:scheduling}

For core $a \in \{S, A\}$ (SNN, ANN), assigning output element $(i,j)$ with $\hat{r}_{i,j}$ effective non-zero matches incurs energy $e_a(i,j)$ and latency $\ell_a(i,j)$:
\begin{equation}
    e_a(i,j) = \epsilon_a \,\hat{r}_{i,j} + \zeta_a, \qquad
    \ell_a(i,j) = \beta_a \,\hat{r}_{i,j} + \delta_a,
    \label{eq:per-element-cost}
\end{equation}
where $(\epsilon_a, \beta_a)$ are the per-match energy and latency coefficients obtained from RTL microbenchmarks, and $(\zeta_a, \delta_a)$ capture fixed per-element overheads. The match count $\hat{r}_{i,j}$ is the 90th-percentile of per-element non-zero intersection counts over a 128-sample calibration set, chosen to guard against tail-latency violations in the makespan.

Let $\mathbf{X} \in \{0,1\}^{M \times N}$ be the assignment matrix for a layer with output dimension $M \times N$, where $\mathbf{X}[i,j]=1$ routes element $(i,j)$ to the SNN core. The total energy sums over all elements:
\begin{equation}
    E(\mathbf{X}) = \textstyle\sum_{(i,j):\,\mathbf{X}[i,j]=1} e_S(i,j) + \sum_{(i,j):\,\mathbf{X}[i,j]=0} e_A(i,j).
    \label{eq:energy}
\end{equation}
Each compute core $a$ with $P_a$ number of PEs has a delay $T_a=\max_{p \leq P_a} \sum_{(i,j) \in \mathcal{S}_{a,p}} \ell_a(i,j)$ and the layer delay is $D(\mathbf{X}) = \max\{T_S, T_A\}$. We minimize:
\begin{equation}
    \text{EDP}(\mathbf{X}) = E(\mathbf{X}) \cdot D(\mathbf{X}).
    \label{eq:edp}
\end{equation}
Energy is separable over elements but the delay couples all assignments through the makespan, making EDP minimization NP-hard. We solve Eq.~\eqref{eq:edp} offline using simulated annealing with element-flip moves. Each move flips a single $\mathbf{X}[i,j]$, re-evaluates the affected core's makespan in $O(P_a)$, and accepts if the EDP decreases or with probability $\exp(-\Delta\text{EDP}/T_{\text{S/A}})$ under a geometric cooling schedule. The initial assignment routes each element to the core with lower marginal cost. The solver runs $10 \cdot M \cdot N$ iterations per layer, completing in under 2 seconds for a $512 \times 512$ output on a AMD EPYC 9555 processor. The resulting per-layer binary mask (mode) is stored alongside the model weights and applied at inference with no runtime overhead.

\section{Experimental Methodology}
\label{sec:methodology}

\subsection{System Configuration}
\label{sec:methodology:system}

We evaluate NeuroFlex using the configuration summarized in Table~\ref{tab:config}. The SNN cores perform spike-based computations over 8 timesteps, which is sufficient to achieve lossless conversion as discussed in Section~\ref{sec:dataflow-mode}. Each core contains 16 PEs sharing a 512~KB double-buffered global cache operated across 32 banks. The microarchitectural details of the ANN and SNN PEs are described in Section~\ref{sec:PE}. We allocate 128~GB/s of off-chip memory bandwidth via 32 $\times$ 64-bit HBM channels, which our analysis confirms is sufficient to sustain on-chip reuse and keep PEs busy across all evaluated workloads.

\begin{table}[t]
\centering
\caption{NeuroFlex configuration.}
\label{tab:config}
\begin{tabular}{l l}
\toprule
\textbf{Component} & \textbf{Setting} \\
\midrule
SNN cores & 16 PEs, 8-bit weight \\
ANN cores & 16 PEs, 8-bit weight \\
Inner-join unit & 32 inner-join units \\
Global cache & 512~KB, 32 banks, 32-way associative \\
Crossbars & $32\times32$ and $32\times32$, swizzle-switch based \\
Main memory & 128~GB/s over 32 $\times$ 64-bit HBM channels \\
\bottomrule
\end{tabular}
\end{table}

\subsection{Software Configuration}
\label{sec:methodology:software}
We evaluate NeuroFlex across seven models spanning three domains: CNNs (VGG-16, ResNet-34, GoogleNet), transformers (BERT, ViT-S), and large language models (GPT-2~XL, OPT-6.7B). Each network is first trained with QCFS activation and then converted to its PASCAL-equivalent SNN, so that both execution modes share bit-identical weights. Because the PASCAL conversion is integer-exact (Section~\ref{sec:dataflow-mode}), the ANN and SNN realizations of each element produce identical outputs, so element-level mode switching incurs no accuracy loss by construction. The two modes are interchangeable realizations of the same integer computation, and the reported metrics apply equally to both. We use quantization step $L=8$ and timestep budget $T=L=8$ throughout, the configuration under which the PASCAL algorithm reproduces the QCFS activation exactly. Table~\ref{tab:accuracy} reports the resulting accuracy and perplexity at $L=8, T=8$.

\begin{table}[t]
\centering
\caption{Accuracy and perplexity of the PASCAL SNN ($T{=} L {=} 8$).}
\label{tab:accuracy}
\begin{tabular}{llc}
\toprule
\textbf{Model} & \textbf{Dataset (Metric)} & \textbf{PASCAL ($T{=}8$)} \\
\midrule
VGG-16     & CIFAR-10 (Acc.\ \%)             & 96.93  \\
GoogleNet  & ImageNet (Acc.\ \%)             & 65.71  \\
ViT-S      & CIFAR-10 (Acc.\ \%)             & 97.01  \\
BERT       & SST-2 (Acc.\ \%)                & 91.86  \\
GPT-2~XL   & WikiText-103 (PPL $\downarrow$)  & 11.16 \\
OPT-6.7B   & WikiText-103 (PPL $\downarrow$)  & 9.31   \\
\bottomrule
\end{tabular}
\end{table}

\subsection{Baselines}
\label{sec:methodology:baselines}

\paragraph{Hardware baselines}
We compare NeuroFlex against two primary hardware baselines. For the ANN baseline, we use SparTen\cite{SparTen}, which employs the same bitmap encoding and inner-product dataflow as NeuroFlex. Since our innovation is on the compute side (element-level hybrid scheduling and PASCAL-based SNN execution), SparTen provides a direct apples-to-apples comparison that isolates our gains from dataflow or encoding differences. For the SNN baseline, we use LoAS~\cite{yin2024loas}, the only dual-sparse SNN accelerator that also employs bitmap encoding, making it the most architecturally comparable SNN design. All baselines and NeuroFlex are configured identically: 32 PEs in total, a 512~KB double-buffered global cache, and a 128~GB/s HBM subsystem. To ensure functional comparability, we replace activations in LoAS with PASCAL's spiking activation and SparTen's activation with QCFS.

\paragraph{Algorithmic baselines}
To isolate the contribution of element-level scheduling from the hardware, we evaluate three scheduling strategies on identical NeuroFlex hardware, varying only the granularity of the assignment. Our EDP-based cost function is applied uniformly at all three granularities, so the comparison isolates granularity alone rather than confounding it with cost-awareness. \emph{LayerWise} assigns all elements in a layer to a single mode, following the coarse-grained strategy of EPHA~\cite{zhao2024epha} and NEBULA~\cite{singh2020nebula}. \emph{TileWise} partitions each output matrix into fixed-size tiles and assigns entire tiles to a core, following C-DNN~\cite{cdnn} and C-Transformer~\cite{c-transformer}. \emph{ElementWise} applies the same cost-guided scheduler at element granularity. Since all three run on identical hardware with the same PE count and the same cost model, any performance difference is attributable entirely to scheduling granularity, forming a clean ablation from coarse to fine-grained assignment and demonstrating that element-level assignment is the superior choice.

\subsection{Implementation and Modeling}
\label{sec:methodology:impl}

We implement key components of NeuroFlex and all baselines in RTL and synthesize with Synopsys Design Compiler at 560~MHz in a 40~nm technology node. We use CACTI~7.0\cite{cacti7} to model the memory structures. We develop a cycle-level simulator in Python that tiles loops and maps them to hardware, modeling execution on NeuroFlex and all baselines under identical kernels, dataflows, and memory hierarchies.

\section{Experimental Results}
\label{sec:results}

We report speedup (higher is better), energy efficiency (higher is better), and Energy-Delay Product (EDP, lower is better). Speedup is normalized to the SNN-only latency. Energy efficiency is normalized to the ANN-only energy. EDP is normalized to SparTen unless noted otherwise.

\subsection{End-to-End Performance}

\begin{figure*}[h!]
  \centering
  \includegraphics[width=1\textwidth]{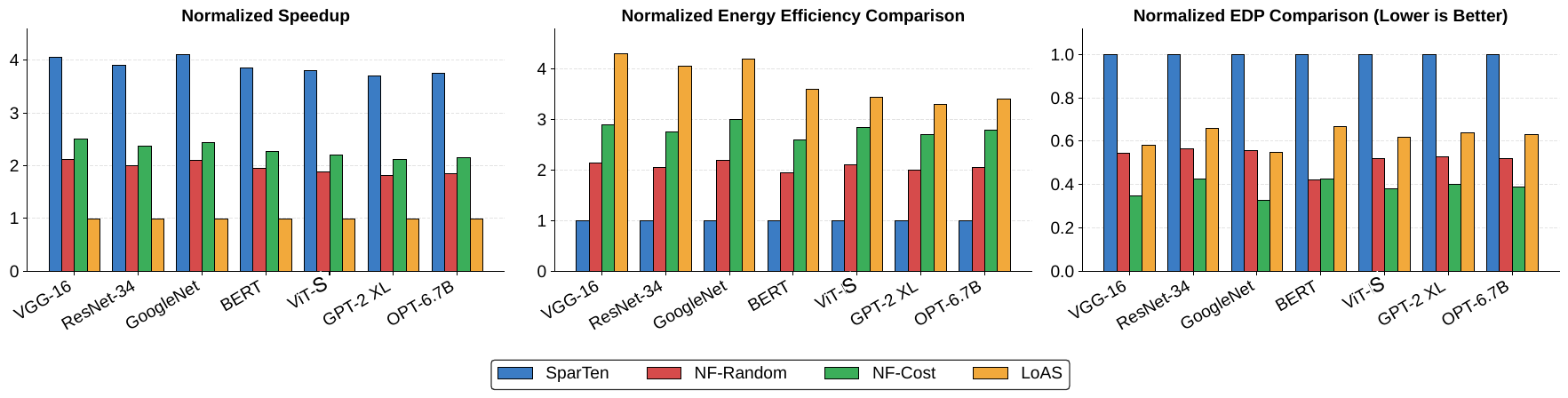}
  \caption{End-to-end speedup and energy-efficiency across baseline SNN and ANN accelerators.}
  \label{fig:speed-energy}
\end{figure*}

Figure~\ref{fig:speed-energy} compares NeuroFlex against SparTen (ANN-only) and LoAS (SNN-only) across seven models spanning CNNs (VGG-16, ResNet-34, GoogleNet), transformers (BERT, ViT-S), and large language models (GPT-2~XL, OPT-6.7B). Since NeuroFlex, SparTen, and LoAS share the same bitmap encoding, FiberCache, and memory hierarchy, the speedup and energy differences reported here isolate \emph{compute-side} gains, specifically, the effect of element-level hybrid scheduling and PASCAL-based SNN execution versus single-mode ANN or SNN compute. The memory subsystem contributes equally to all three designs and cancels out in the normalized comparison.

NeuroFlex with its cost-based scheduling improves throughput over random element assignment by 17.8\% for VGG-16, 18.9\% for ResNet-34, 16.2\% for GoogleNet, and 16.7\% for BERT. The gains stem from the per-element placement decisions that match each output element to the core whose marginal energy-delay cost is the lowest, together with cost-guided scheduler packing that reduces the inter-core makespan imbalance. Energy efficiency also improves for three of the four core models: routing high-sparsity elements to the SNN core replaces the MAC operations with cheaper accumulate-only arithmetic, lowering dynamic energy without affecting throughput on the remaining ANN elements. SparTen remains the fastest single-mode baseline because it avoids the SNN temporal overhead entirely, but NeuroFlex closes the speed gap while delivering substantially higher energy efficiency than ANN-only execution.

LoAS achieves the highest energy efficiency among all configurations because every element executes with accumulate-only arithmetic and binary spike encoding. However, the temporal overhead of processing timesteps per element inflates its latency, leaving its EDP significantly worse than NeuroFlex despite the energy advantage.

\subsection{Energy-Delay Product}

\begin{figure*}[h!]
  \centering
  \includegraphics[width=1.0\textwidth]{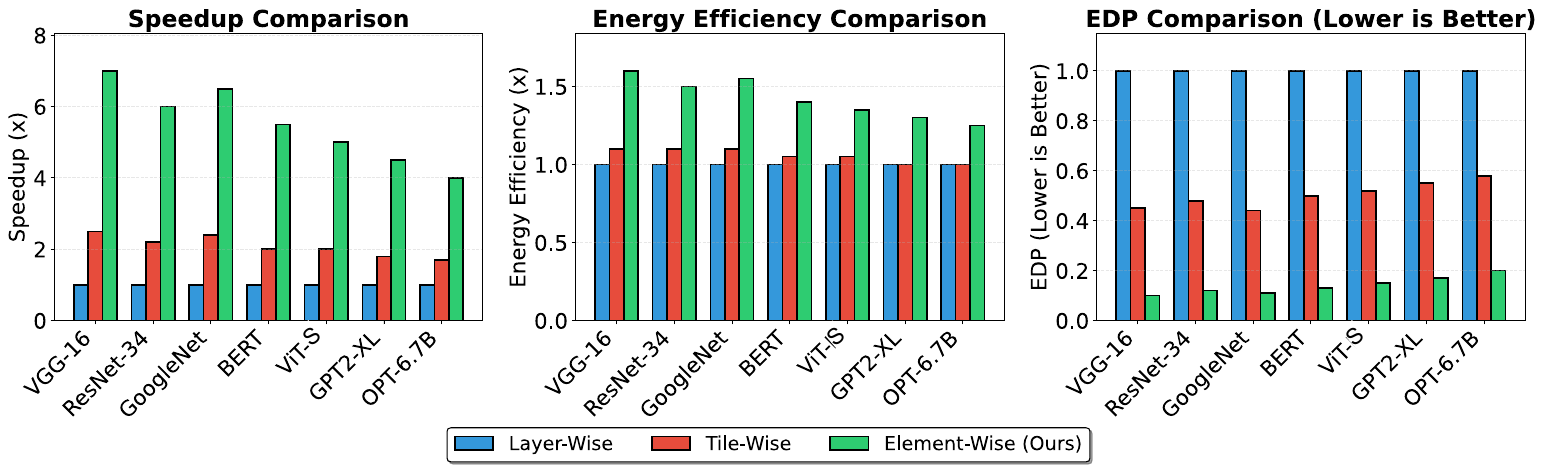}
  \caption{Speedup, Energy, and EDP comparison across different mapping strategies - Layerwise, TileWise, Elementwise.}
  \label{fig:edp}
\end{figure*}

Figure~\ref{fig:speed-energy} combines latency and energy into the energy-delay product, where a lower value indicates better overall efficiency. NeuroFlex with cost-based scheduling reduces EDP versus SparTen by 65\% for VGG-16, 57.5\% for ResNet-34, 67.1\% for GoogleNet, and 57.4\% for BERT. Random element assignment also improves over single-mode baselines but trails the cost-based policy, providing EDP reductions of 45.4\%, 43.4\%, 44.4\%, and 57.6\% for the respective models. The gap between random and cost-based scheduling quantifies the value of the surrogate-based scorer: random assignment captures the benefit of element-level granularity but misses the per-element energy-delay matching that the cost function provides.

The SNN-only baseline (LoAS) achieves the lowest absolute energy, but its longer delay yields 57.1-69.5\% higher EDP than NeuroFlex. This confirms that energy efficiency alone is insufficient, the product of energy and delay must be jointly minimized, which is precisely what NeuroFlex's cost-guided scheduler targets.

\subsection{Scheduling Granularity Ablation}

Figure~\ref{fig:edp} also serves as an ablation over scheduling granularity, since all three strategies, layer-wise, tile-wise, and cost-based element-wise, execute on identical NeuroFlex hardware with the same PE count. Layer-wise scheduling assigns every element in a layer to a single mode. When a layer runs in SNN mode, all ANN cores sit idle, and vice versa. This mutual exclusion caps PE utilization at 60-76\% (Figure~\ref{fig:pe-util}) and inflates both latency and EDP. Tile-wise scheduling partitions each output matrix into fixed-size tiles and assigns entire tiles to a core. While this captures some intra-layer variation, tiles are coarser than individual elements: a tile containing a mix of dense and sparse elements must commit to one mode, wasting energy on the dense elements if mapped to SNN or forgoing the energy savings of the sparse elements if mapped to ANN. Tile boundaries also introduce alignment and padding overhead. As a result, tile-wise scheduling improves over layer-wise but still under-utilizes the fabric compared to element-level execution.

Element-wise scheduling eliminates all intra-group heterogeneity: every element is independently matched to its optimal mode. Cost-based scoring further refines this by using the surrogate objective to rank elements by their marginal energy-delay trade-off, then packing them onto PEs with cost-guided scheduler to minimize makespan. The result is 97-99\% PE utilization and the lowest EDP across all models.

\subsection{Comparison with Gamma and Prosperity}

\begin{figure}[h!]
  \centering
  \includegraphics[width=0.9\linewidth,height=0.5\textheight,keepaspectratio]{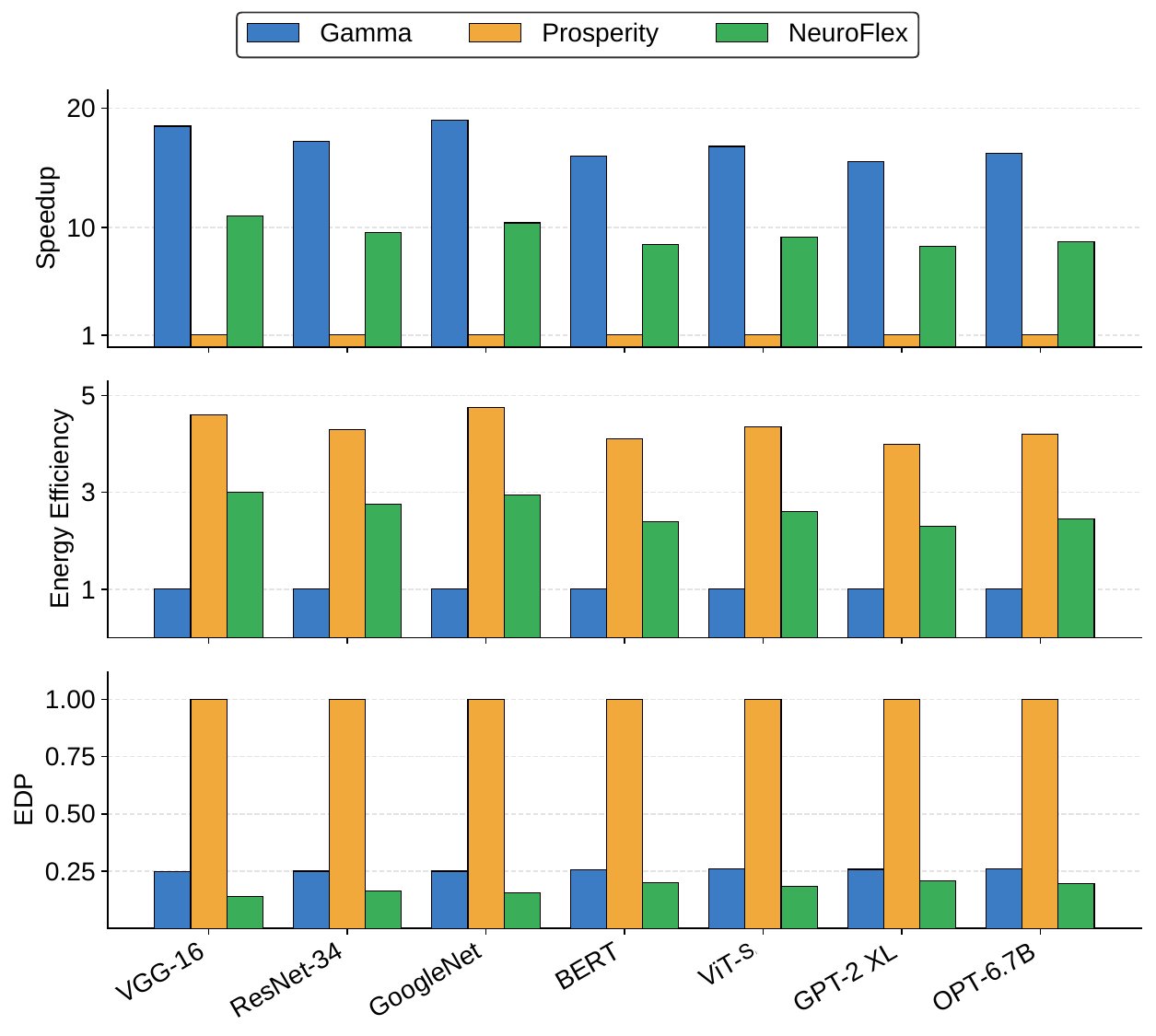}
  \caption{Comparison of NeuroFlex with Gamma and Prosperity. (a)~Speedup, (b)~Energy Efficiency, (c)~Energy-Delay Product.}
  \label{fig:prosperity}
\end{figure}

Figure~\ref{fig:prosperity} compares our NeuroFlex against Gamma \cite{gamma} and Prosperity \cite{wei2025prosperityacceleratingspikingneural} on an \emph{end-to-end} basis, accounting for both computational and memory energy. Gamma and Prosperity employ fundamentally different sparsity encodings, CSR-style coordinates and product-sparsity metadata, respectively, which significantly impact data movement. An end-to-end evaluation is therefore necessary to capture these effects.
Gamma accelerates sparse matrix multiplication by relying on a CSR-style coordinate encoding, which incurs multi-byte metadata per non-zero element. The cost of fetching the required metadata dominates Gamma's energy consumption, resulting in lower energy efficiency despite providing high compute throughput. As illustrated in Figure~\ref{fig:prosperity}, Gamma achieves the highest speedup (up to $\sim$19$\times$), but its energy efficiency remains at the baseline.
Prosperity exploits product sparsity, skipping entire output elements whose dot products evaluate to zero. This is a coarse-grained optimization: an output is skipped only if \emph{all} activation-weight pairs contributing to it contain at least one zero operand. In contrast, NeuroFlex exploits dual sparsity, skipping zero-valued operand pairs within each dot product. This finer-grained approach eliminates unnecessary computation even when the output is non-zero, leading to significantly higher effective utilization.
As a result, NeuroFlex achieves substantial speedups over Prosperity (e.g., $\sim$12$\times$ on VGG-16) while also improving energy efficiency (2.3-3.0$\times$). Although Gamma attains higher raw speedup, its relative lack of energy efficiency limits its overall benefit. When considering energy-delay product (EDP), NeuroFlex provides the best trade-off: it reduces EDP by approximately $5$-$6\times$ relative to Prosperity and by about $1.5$-$2\times$ compared to Gamma across models.
Overall, these results demonstrate that fine-grained dual-sparsity exploitation in NeuroFlex achieves a superior balance of performance and energy efficiency compared to both CSR-based acceleration (Gamma) and coarse-grained product sparsity (Prosperity).

\subsection{PE Utilization}

\begin{figure}[h!]
  \centering
  \includegraphics[width=0.5\textwidth]{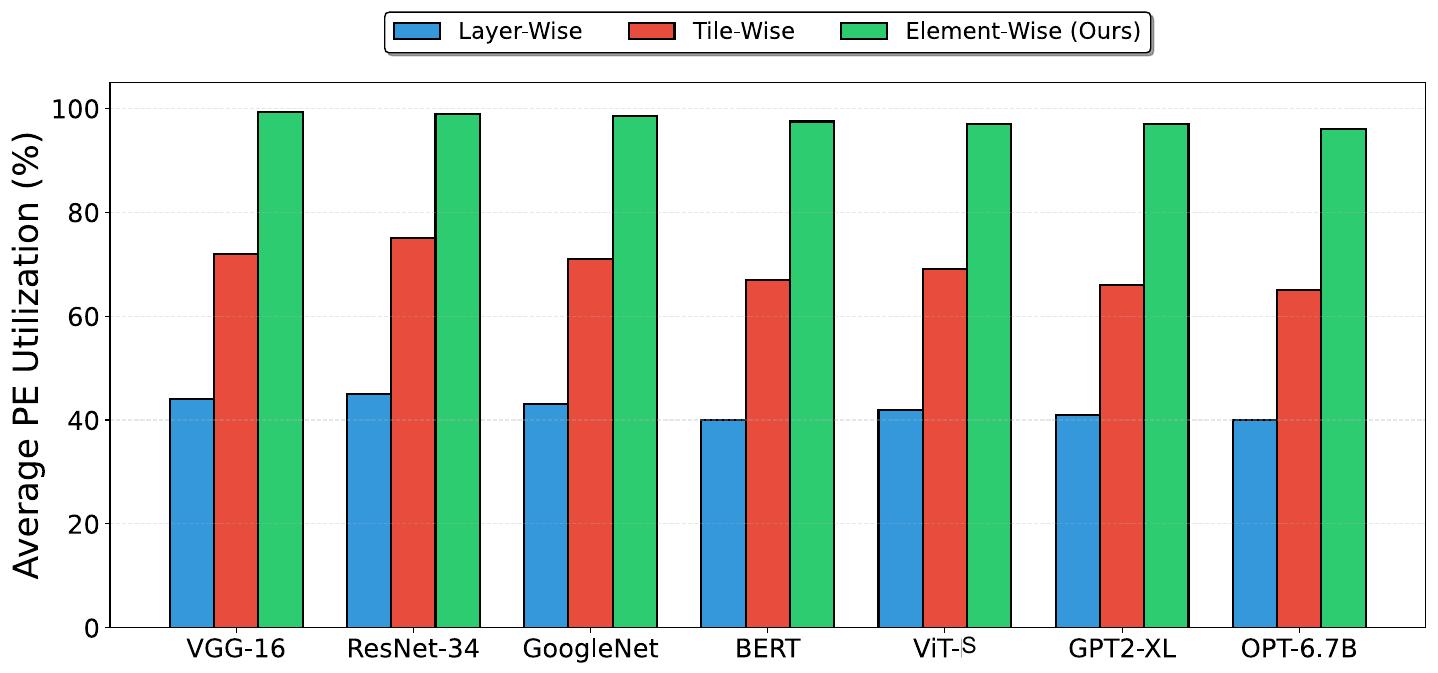}
  \caption{Average PE utilization across models and mapping strategies.}
  \label{fig:pe-util}
\end{figure}

Figure~\ref{fig:pe-util} quantifies how scheduling granularity affects array utilization. Cost-based element-wise scheduling saturates the PE array: VGG-16, ResNet-34, and GoogleNet achieve 99.3\%, 98.9\%, and 98.6\% average utilization (calculated as \cite{derose2007detecting}), respectively, while BERT attains 97.5\%. The near-perfect utilization arises because the scheduler distributes fine-grained work across PEs, leaving negligible idle slots.


Tile-wise hybridization achieves only 71.0-75.9\% utilization on VGG-16, 80.9-83.9\% on ResNet-34, 70.7-73.7\% on GoogleNet, and 60.3-63.9\% on BERT. The low utilization reflects the fundamental limitation of tile-level mode switching: whichever core type is not active for the current layer contributes zero useful work, creating large idle windows that coarse scheduling cannot fill. These results confirm that element-level granularity is necessary to keep both ANN and SNN cores consistently busy.

\subsection{Scalability}

\begin{figure}[h!]
  \centering
  \includegraphics[width=0.45\textwidth, clip]{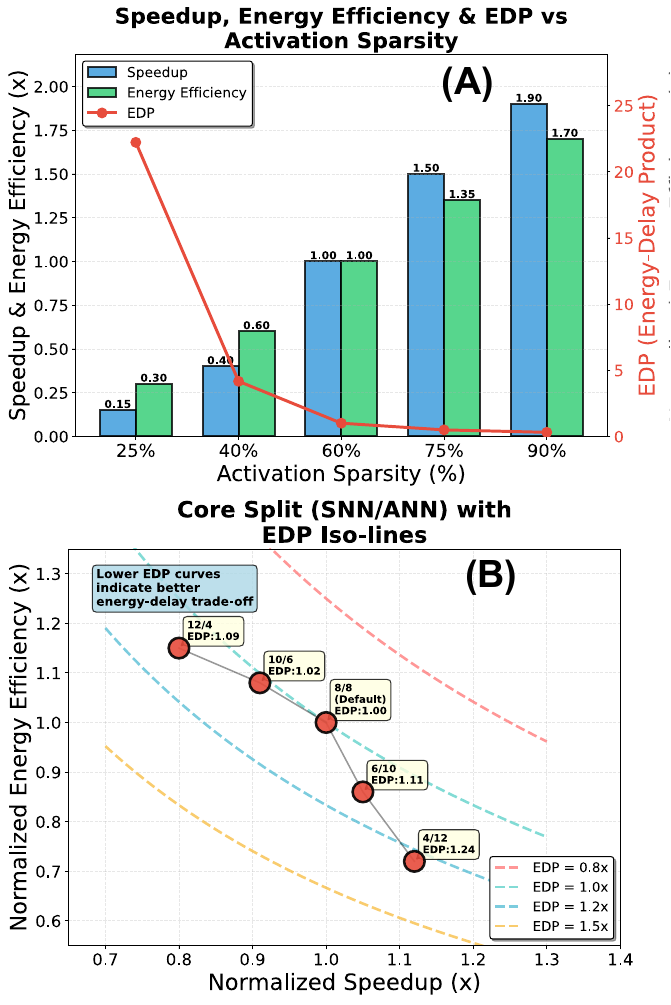}
  \caption{Scalability. \textbf{(A)} Speedup, energy efficiency, and EDP
    as a function of activation sparsity, swept from 25\% to 90\%. Speedup
    and EDP both scale with the fraction of activation-weight pairs skipped
    by the bitmap inner-join, reflecting the sparsity-dependence of
    NeuroFlex's dual-sparsity exploitation. \textbf{(B)} Normalized speedup
    versus energy efficiency across five SNN/ANN core split ratios (4/12,
    6/10, 8/8, 10/6, 12/4), with dashed EDP iso-lines overlaid to visualize
    the energy-delay trade-off surface. Points closer to the lower-left
    iso-lines represent better EDP; the 8/8 balanced split sits nearest the
    EDP knee.}
  \label{fig:scalability}
\end{figure}

Figure~\ref{fig:scalability} investigates how NeuroFlex's performance scales along the following axes.
\paragraph{Activation sparsity (Panel~A)}
As activation sparsity decreases from 90\% to 60\% to 25\%, relative speedup drops from 100\% to 21.3\% to 12.5\%. This trend directly reflects the bitmap inner-join mechanism: at high sparsity, the majority of activation-weight pairs are matched-zero and skipped, so the PE processes only a small fraction of the full dot product. As sparsity decreases, more pairs survive the inner-join, increasing the effective workload per element and reducing the relative advantage of dual-sparsity exploitation. The EDP also rises with decreasing sparsity, confirming that NeuroFlex's efficiency gains are proportional to the sparsity available in the workload.

\paragraph{Core split (Panel~B)}
The allocation of cores between ANN and SNN modes tunes the energy-delay trade-off along a Pareto front. Allocating more ANN cores (SNN/ANN\,=\,6/10) raises speedup to 1.12$\times$ but reduces energy efficiency to 86\% of the balanced configuration, because the faster ANN MACs consume more power per operation. Conversely, allocating more SNN cores (SNN/ANN\,=\,10/8) raises energy efficiency to 1.15$\times$ at 91\% speed, since more elements execute via the cheaper accumulate-only path. The balanced 8/8 split sits near the EDP knee, confirming it as the optimal configuration.

Extreme splits reveal diminishing returns: at SNN/ANN = 12/4, energy efficiency saturates because few elements remain that benefit from SNN execution, while the reduced ANN core count bottlenecks latency. At SNN/ANN\,=\,4/12, speed plateaus because most high-match elements are already on the ANN cores, and the excess ANN PEs sit partially idle. The Pareto front in Panel~B provides system designers with a principled method to select the core split based on their deployment constraint, whether latency-sensitive (favor more ANN cores) or energy-constrained (favor more SNN cores).

\subsection{Power Analysis}

\begin{figure}[t]
  \centering
  \includegraphics[width=0.40\textwidth, trim = 0cm 0cm 2cm 0cm ]{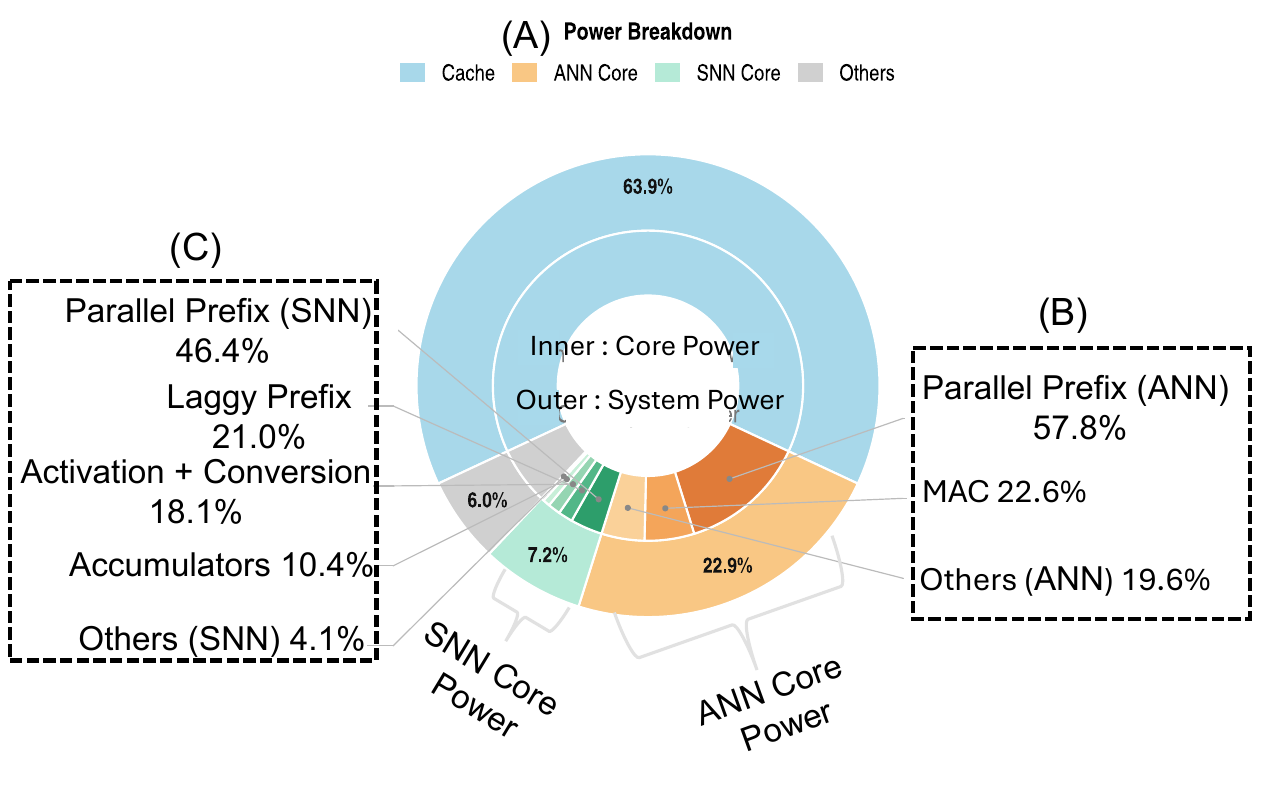}
  \caption{Power analysis. \textbf{A}: System-level power breakdown.
  \textbf{B}: ANN core internal breakdown. \textbf{C}: SNN core internal breakdown.}
  \label{fig:power}
\end{figure}

Figure~\ref{fig:power}A shows that the global cache dominates system power at 63.9\%, followed by the ANN cores (28.88\%) and SNN cores (7.22\%). The architectural implication is clear: optimizing on-chip data reuse and bank-level concurrency within the FiberCache provides larger system-level gains than micro-optimizing individual compute units.

Within the ANN core (Figure~\ref{fig:power}B), the parallel-prefix logic for dual-sparsity inner-join is the primary power consumer at 59\%, with MAC units contributing 23.1\% and control/other logic at 19.9\%. The prefix dominance reflects the cost of fast bitmap intersection: the one-cycle prefix circuit that enables single-cycle address alignment for each chunk requires a wide parallel tree that draws significant switching power. Within the SNN core (Figure~\ref{fig:power}C), parallel-prefix consumes 44.4\%, the laggy prefix contributes 17.3\%, activation and conversion logic accounts for 10\%, accumulators 3.9\%, and other logic 20.1\%. The laggy prefix, a secondary accumulation circuit that corrects for the latency difference between the fast and laggy prefix paths, adds power overhead unique to the SNN pipeline but is necessary for cycle-accurate partial-sum updates.

These distributions confirm that dual-sparsity exploitation via bitmap inner-join is not free in terms of power, but the compute cores collectively account for only 36.1\% of system power. Memory traffic through the cache hierarchy dominates, which is consistent with the broader trend in sparse accelerator design where data movement, not computation, is the primary energy bottleneck.

\paragraph{Activation operators}
NeuroFlex uses QCFS in the ANN core and the PASCAL integrate-fire-reset sequence in the SNN core, both of which are more expensive than the conventional ReLU and LIF units they replace: ReLU costs $0.8\times$ the power of QCFS, and LIF costs $0.4\times$ the power of PASCAL. Despite this overhead, both activation engines together contribute to $<1\%$ of total system power, so the price of lossless integer-exact conversion is negligible at the system level. Combined with the power breakdown above, this indicates that NeuroFlex's cost-based scheduling reduces the time component of EDP without meaningful activation overhead, and that future optimization effort is best directed at the memory hierarchy rather than the compute pipeline.

\section{Conclusion}
\label{sec:conclusion}
In this work, we proposed the NeuroFlex accelerator for fine-grained and integer-exact co-execution of ANNs and SNNs at the element level, which lowers energy-latency cost on sparse workloads while preserving accuracy. The design extends QCFS-based ANN-SNN conversion to independent elements, unifies storage with on-the-fly spike generation, and uses an offline cost model to keep ANN and SNN cores busy without runtime overhead. Across all the different kinds of workloads, our scheduler improved throughput and reduced EDP versus strong ANN-only and SNN-only baselines. The key limitation is scope: our approach is valid only for models whose activations can be represented with QCFS. Models that require non-QCFS activations are outside our guarantees. Future work can broaden the activation set, explore workload-aware scheduling, and validate at larger system scales.

\section*{Acknowledgment}
 We thank the Robert Bosch Centre for Data Science and AI (RBCDSAI) and the Centre for Development of Advanced Computing (C-DAC) for providing GPU computing resources. We also thank Rajshekhar Rakshit, Rakshitha Kalkura, and Preethi Carmel Bosco of the BrainSeek Lab, Department of Computer Science and Engineering, IIT Madras, for training, fine-tuning, and running inference on the models reported in this work.

 \newpage

\appendix
\section*{Artifact Appendix}

\subsection{Abstract}

Artifact evaluation focuses on the accuracy and perplexity of PASCAL SNN, discussed in Section~\ref{sec:methodology:software} and in Table~\ref{tab:accuracy}. All six evaluations run at $T{=}8$, $L{=}8$. The code required to reproduce the results presented in the paper can be found at \url{https://github.com/BrainSeek-Lab/NeuroFlex_Code_weights}.

\subsection{Artifact check-list (meta-information)}

{\small
\begin{itemize}
  \item {\bf Algorithm: } PASCAL ANN-SNN conversion, INT8 quantization
  \item {\bf Program: } Python, PyTorch, HuggingFace Transformers, timm
  \item {\bf Model: } VGG-16, GoogleNet, ViT-S, BERT, GPT-2 XL, OPT-6.7B (all
        checkpoints downloaded by the provided script)
  \item {\bf Data set: } CIFAR-10, SST-2, WikiText-103 (downloadable);
        ImageNet (must be supplied by the evaluator)
  \item {\bf Run-time environment: } Linux, CUDA 11.8 or 12.1, conda
  \item {\bf Hardware: } x86\_64 machine. GPU: NVIDIA RTX 6000 Ada or NVIDIA
        H200. At least 24\,GB VRAM recommended
  \item {\bf Execution: } Bash scripts
  \item {\bf Metrics: } Classification accuracy, perplexity
  \item {\bf Output: } Table~\ref{tab:accuracy}
  \item {\bf Experiments: } PASCAL SNN inference at $T = L = 8$ on CNN,
        transformer and LLM workloads
  \item {\bf How much disk space required (approximately)?: } 35\,GB (excluding
        ImageNet)
  \item {\bf How much time is needed to prepare workflow (approximately)?: }
        60 minutes
  \item {\bf How much time is needed to complete experiments (approximately)?: }
        3 hours (H200); 2.5 hours without ImageNet
  \item {\bf Publicly available?: } Yes
  \item {\bf Code licenses (if publicly available)?: } CC BY 4.0
  \item {\bf Workflow automation framework used?: } Bash scripts
  \item {\bf Archived (provide DOI)?: }
        \url{https://doi.org/10.5281/zenodo.21546571}
\end{itemize}
}

\subsection{Description}

\subsubsection{How to access}

Clone the source code from
\url{https://github.com/BrainSeek-Lab/NeuroFlex_Code_weights.git}. Then create
the two conda environments and run the download scripts as described in the
\texttt{README.md} file at the repository root. An archived snapshot of the
evaluated revision (tag \texttt{v1.0-micro-ae}) is available at
\url{https://doi.org/10.5281/zenodo.21546571}.

\subsubsection{Hardware dependencies}

The code has been tested on an x86\_64 machine using NVIDIA RTX 6000 Ada and
H200 GPUs. All experiments require a GPU. We recommend at least 24\,GB of VRAM;
OPT-6.7B requires roughly 16\,GB in fp16 and can be run at a reduced batch size
(\texttt{OPT\_BATCH=1}) or skipped (\texttt{SKIP\_OPT=1}) on smaller cards.

\subsubsection{Software dependencies}

Two conda environments are required, as the vision and NLP workloads depend on
incompatible PyTorch and \texttt{transformers} versions. The vision environment
uses \texttt{torch 2.7.1+cu118}, \texttt{torchvision 0.22.1+cu118} and
\texttt{timm 1.0.28}; the NLP environment uses \texttt{torch 2.5.1} and
\texttt{transformers 4.40.0}. Python 3.10 and CUDA 11.8+ are required. The
creation commands are listed in the \texttt{README.md} file, and the two
interpreters are passed to the run scripts through the
\texttt{PYTHON\_VISION} and \texttt{PYTHON\_NLP} environment variables.

\subsubsection{Data sets}

CIFAR-10, SST-2 and WikiText-103 are downloaded by
\texttt{scripts/download\_datasets.sh}. ImageNet is not downloadable, as it
requires a signed licence agreement; evaluators who have a copy should place it
in \texttt{ImageFolder} layout and export \texttt{NEUROFLEX\_IMAGENET}.
ImageNet is used by one row of Table~\ref{tab:accuracy}, which is skipped automatically
when it is absent.

\subsubsection{Models}

Users do not need to download the models separately. All six PASCAL SNN
checkpoints are fetched by \texttt{scripts/download\_checkpoints.sh}. The
exception is BERT, for which only the full-parameter fine-tuned checkpoint is
released; \texttt{run\_transformers.sh} therefore fine-tunes on SST-2 before
evaluating, which takes $\approx$20 minutes.

\subsection{Installation}

After cloning the GitHub repository and creating the two environments, export
the interpreters and fetch the weights and datasets:

{\footnotesize
\begin{verbatim}
export PYTHON_VISION=/path/to/vision-env/bin/python
export PYTHON_NLP=/path/to/nlp-env/bin/python
PYTHON=$PYTHON_VISION \
  bash scripts/download_checkpoints.sh
bash scripts/download_datasets.sh
\end{verbatim}
}

If ImageNet is available, additionally export its location:

{\footnotesize
\begin{verbatim}
export NEUROFLEX_IMAGENET=/path/to/imagenet
\end{verbatim}
}

\subsection{Experiment workflow}

We use bash scripts to automate the workflow for generating results for
artifact evaluation. To reproduce Table~\ref{tab:accuracy}, simply execute:

\begin{verbatim}
/bin/bash scripts/run_all.sh
\end{verbatim}

If ImageNet is not available, execute the following instead, which runs all
rows except GoogleNet:

\begin{verbatim}
/bin/bash scripts/run_all_no_imagenet.sh
\end{verbatim}

Individual model categories (shown in Table~\ref{tab:accuracy}) can also be executed separately using \texttt{scripts/run\_cnns.sh}, \texttt{scripts/run\_transformers.sh} and \texttt{scripts/run\_llms.sh}. After the scripts complete, the results will be available in the \texttt{results/} sub-directory. Additional details can be found in the \texttt{README.md} file.

\subsection{Evaluation and expected results}

After running the scripts as described above, the generated log files will be
available locally in the \texttt{results/} sub-directory, with the accuracy or
perplexity of each model printed at the end of its run. These values correspond
to Table~\ref{tab:accuracy} in the original paper: VGG-16 on CIFAR-10 (96.93), GoogleNet on
ImageNet (65.71), ViT-S on CIFAR-10 (97.01), BERT on SST-2 (91.86), GPT-2 XL on
WikiText-103 (11.162 PPL) and OPT-6.7B on WikiText-103 (9.31 PPL). Any row
whose checkpoint or dataset is missing is skipped with a message rather than
aborting the run. Additional details are provided in the \texttt{README.md}
file under the section {\em Reproduction notes}.

\subsection{Experiment customization}

The timestep and quantization step are exposed through the
\texttt{NEUROFLEX\_T} and \texttt{NEUROFLEX\_L} environment variables, both of
which default to 8. Per-row batch sizes, the data, checkpoint and result paths,
and flags to skip individual LLM rows are also configurable; the complete list
is given in the \texttt{README.md} file.

\subsection{Notes}

Some variation in the reproduced accuracy and perplexity may occur, as the
results depend on the random seed, the GPU type, the CUDA and cuDNN versions,
and non-deterministic kernel selection. However, the overall trends will remain
consistent with those presented in the paper.

\subsection{Methodology}

Submission, reviewing and badging methodology:

\begin{itemize}
  \item \url{https://www.acm.org/publications/policies/artifact-review-and-badging-current}
  \item \url{https://cTuning.org/ae}
\end{itemize}

\bibliographystyle{IEEEtran}
\bibliography{sample-base}

\end{document}